\pdfoutput=1
\documentclass[11pt]{article}

\usepackage[preprint]{acl}

\usepackage{times}
\usepackage{latexsym}

\usepackage[T1]{fontenc}
\usepackage[utf8]{inputenc}

\usepackage{microtype}

\usepackage{inconsolata}

\usepackage{graphicx}

\usepackage{subcaption}
\usepackage{amsmath}
\usepackage{amsfonts}
\usepackage{amssymb}
\usepackage{algorithm}
\usepackage{algorithmic}

\usepackage{color}
\definecolor{Red}{RGB}{255,0,0}
\definecolor{Green}{RGB}{0,176,80}
\definecolor{Blue}{RGB}{0,176,240}
\definecolor{Purple}{RGB}{112,048,160}

\usepackage{tcolorbox}
\tcbuselibrary{breakable}

\title{Chain-of-Thought Tokens are Computer Program Variables}

\author{Fangwei Zhu, Peiyi Wang, Zhifang Sui \\
  School of Computer Science, \\
  State Key Laboratory of Multimedia Information Processing, \\ 
  Peking University \\
  \texttt{zhufangwei2022@stu.pku.edu.cn} \\
  \texttt{wangpeiyi9979@gmail.com} \\
  \texttt{szf@pku.edu.cn} \\
}

\begin{document}
\maketitle
\begin{abstract}
Chain-of-thoughts (CoT) requires large language models (LLMs) to generate intermediate steps before reaching the final answer, and has been proven effective to help LLMs solve complex reasoning tasks.
However, the inner mechanism of CoT still remains largely unclear.
In this paper, we empirically study the role of CoT tokens in LLMs on two compositional tasks: multi-digit multiplication and dynamic programming.
While CoT is essential for solving these problems, we find that preserving only tokens that store intermediate results would achieve comparable performance.
Furthermore, we observe that storing intermediate results in an alternative latent form will not affect model performance.
We also randomly intervene some values in CoT, and notice that subsequent CoT tokens and the final answer would change correspondingly.
These findings suggest that CoT tokens may function like variables in computer programs but with potential drawbacks like unintended shortcuts and computational complexity limits between tokens.
The code and data are available at \url{https://github.com/solitaryzero/CoTs_are_Variables}.
\end{abstract}

\section{Introduction}
Chain-of-thoughts (CoT)~\cite{wei2022chain} is a widely adopted technique that greatly boosts the capability of large language models (LLMs) in reasoning tasks like solving mathematical problems~\cite{shao2024deepseekmath, wang2024math} or generating codes~\cite{guo2025deepseek}.
By requiring language models to generate intermediate steps before reaching the final result, chain-of-thoughts enables LLMs to perform advanced reasoning, and thus significantly outperforms standard supervised learning methods.
Various methods have been explored to unlock the ability of chain-of-thought reasoning in LLMs, for example designing prompts~\cite{wei2022chain, khotdecomposed, zhouleast}, instruction tuning~\cite{yuemammoth, yumetamath} or reinforcement learning~\cite{havrilla2024teaching, wang2024math, guo2025deepseek}.

There have been theoretical studies on the efficacy of chain-of-thoughts~\cite{deng2024explicit, lichain, chen2024theoretical}.
These studies reveal that while it is exponentially difficult for language models to solve compositional problems requiring serial computations, CoT could help models solve problems under multinominal complexity.
More interestingly, CoT tokens do not need to fall in the ``language space'', using latent vectors could also enable language models to perform complex reasoning~\cite{hao2024training}, indicating that CoTs are more than mere thought traces.
% The mechanism of how CoT works, and the role of CoT tokens still remain underexplored. 
The mechanism of how CoT works, and the role of CoT tokens are still not fully explored. 

% In this paper, we empirically study the role that CoT tokens play in the process of solving problems, and propose the hypothesis that \textbf{CoT tokens function like computer program variables}. 
% Specifically, the tokens in CoT store intermediate values that will be used in subsequent computations, and these values are partially mutable to control the final output.

In this paper, we propose a novel hypothesis that \textbf{CoT tokens function like computer program variables}.
To be specific, the tokens in CoT store intermediate values that will be used in subsequent computations, and these values are partially mutable to control the final output.
As long as the important intermediate values are calculated and stored, the CoT that leads to the final answer could be represented in different forms. 

To verify the hypothesis, we conduct empirical study on two types of problems that both require long-chain serial computations: multi-digit multiplication and dynamic programming.
% By comparing the performance of vanilla prompting with CoT prompting, we confirm that chain-of-thoughts could enable language models to solve previously unsolvable problems.
By comparing the performance of vanilla prompting with CoT prompting, we confirm that CoT is crucial for these problems.
However, we also find that removing non-result tokens would not bring performance drops, which means that tokens storing intermediate values matter more in chain-of-thoughts.

We further explore whether intermediate values could be represented in different forms.
% We find that we can effectively reduce the length of CoT by simply compressing consequent number digits with a single vector, which indicates that the existence, rather than the form, of crucial intermediate variables is what matters most to language models.
We attempt to compress consequent number digits within a single latent vector, and experimental results show that it does not detriment the model performance.
This phenomenon indicates that the existence, rather than the form, of intermediate values matters more to language models.
% By compressing multiple variables into a single latent token, we can effectively control the scale of CoT.
% However, when the degree of compression exceeds language models' capacity limits of processing serial computation, it would lead to failure in reasoning.
However, when the degree of compression exceeds a certain limit of language models' capacity, it would lead to failure in reasoning.

To further confirm that the intermediate values are causally connected with the output, we intervene in some tokens in CoT, replacing them with random values.
% It can be observed that the outputs of LLMs are generally consistent with expected: LLMs will ignore previous steps, and use the intervened value to perform subsequent computations.
% This phenomenon supports the hypothesis that CoT tokens are causally related with the final result.
It can be observed that LLMs will ignore previous steps, and use the intervened value to perform subsequent computations, supporting that CoT tokens are causally related with the final result.
We conclude that CoT tokens function like the variables in computer programs.

To sum up, in this paper we empirically study the function of CoT tokens, and find that: 
% (1) CoT tokens are crucial for compositional problems requiring long-chain serial computations as they act as variables storing intermediate values; 
(1) The role of CoT tokens is similar to variables in computer programs as they store intermediate values used in subsequent computations; 
(2) The intermediate values could be stored in CoT tokens with different forms;
(3) The values in CoT tokens are causally related to the final output and could be intervened like program variables.
These findings are helpful in understanding alternative forms of CoT, and could assist in designing more concise CoTs.

\section{Preliminary}
\label{sec:preliminary}
\subsection{Chain-of-Thoughts}
Chain-of-thoughts (CoT)~\cite{wei2022chain} is a technique commonly used in decoder-only transformers.
Given the input text $x$, CoT attempts to generate intermediate steps $z$ prior to the final answer $y$.
In other words, instead of modeling the probability distribution $P(y|x)$, CoT attempts to model the joint distribution $P(y,z|x) = P(z|x)P(y|x,z)$.

For convenience, we use two special tokens \texttt{<COT>} and \texttt{</COT>} to separate CoT tokens from the final result in our experiments.

% \subsection{Choosing Compositional Tasks}
\subsection{Compositional Tasks}
It has been noticed that LLMs may fail on seemingly trivial problems like multi-digit multiplication.
The commonality of these problems is that they need strict multi-hop reasoning to derive correct predictions, which requires language models to perform step-to-step reasoning like human intelligence.
In this paper, we choose two representative tasks to study the role of CoT tokens:

\begin{algorithm}
\caption{Digit-wise multiplication}
\label{alg:multi}
\begin{algorithmic}
    \REQUIRE Integer $a$ and $b$
    \ENSURE Value of $a*b$
    \STATE Partial = [~]
    \FOR{Digit $b[i]$ in $b$}
        \STATE carry $\leftarrow$ 0
        \FOR{Digit $a[i]$ in $a$}
            \STATE x $\leftarrow$ $a[i]*b[i]$ + carry
            \STATE digit $\leftarrow$ x/10
            \STATE carry $\leftarrow$ x mod 10
        \ENDFOR
        \STATE res $\leftarrow$ Combine digits and last carry
        \STATE Add res to Partial
    \ENDFOR
    \WHILE{Len(Partial) > 1}
        \STATE x $\leftarrow$ Partial[0] + Partial[1]
        \STATE Partial $\leftarrow$ [x] + Partial[2:]
    \ENDWHILE
    \RETURN Partial[0]
\end{algorithmic}
\end{algorithm}

\paragraph{Multi-digit Multiplication}
Calculating the multiplication result of two multi-digit numbers $(x, y)$ requires executing multiple operations based on procedural rules~\cite{dziri2023faith}.
A commonly adopted solution is the long-form multiplication algorithm, which iteratively calculates the digit-wise multiplication result and adds them up to get the final result.
We describe the algorithm in Algorithm \ref{alg:multi}, see Appendix \ref{sec:appendix_multi} for prompt and dataset construction details.

\begin{algorithm}
\caption{Maximum path sum in a grid}
\label{alg:dp}
\begin{algorithmic}
    \REQUIRE A $m*n$ matrix $W$
    \ENSURE Maximum weight sum $s$ on path
    \STATE DP $\leftarrow$ $m*n$ matrix filled with 0
    \FOR{$i$ in range($m$)}
        \FOR{$j$ in range($n$)}
            \STATE DP[$i$][$j$] = $\max$(DP[$i-1$][$j$], DP[$i$][$j-1$]) + W[$i$][$j$]
        \ENDFOR
    \ENDFOR
    \RETURN DP[$m-1$][$n-1$]
\end{algorithmic}
\end{algorithm}

\paragraph{Dynamic Programming}
Dynamic programming (DP) is an algorithmic paradigm that breaks down complicated problems into simpler sub-problems, and then recursively solves these sub-problems.
In our experiments, we use the ``Maximum Path Sum in a Grid'' problem: \textit{Given a $m \times n$ grid filled with non-negative numbers where only moving downward and rightward is allowed, find a path from top left to bottom right which maximizes the sum of all numbers along its path}.
This is a classic problem that can be solved with dynamic programming in $O(m \times n)$ time.
We describe the algorithm in Algorithm \ref{alg:dp}, see Appendix \ref{sec:appendix_dp} for prompt and dataset construction details.

\begin{figure*}[htbp]
    \centering
    \begin{subfigure}[b]{0.23\textwidth}
        \includegraphics[width=\linewidth]{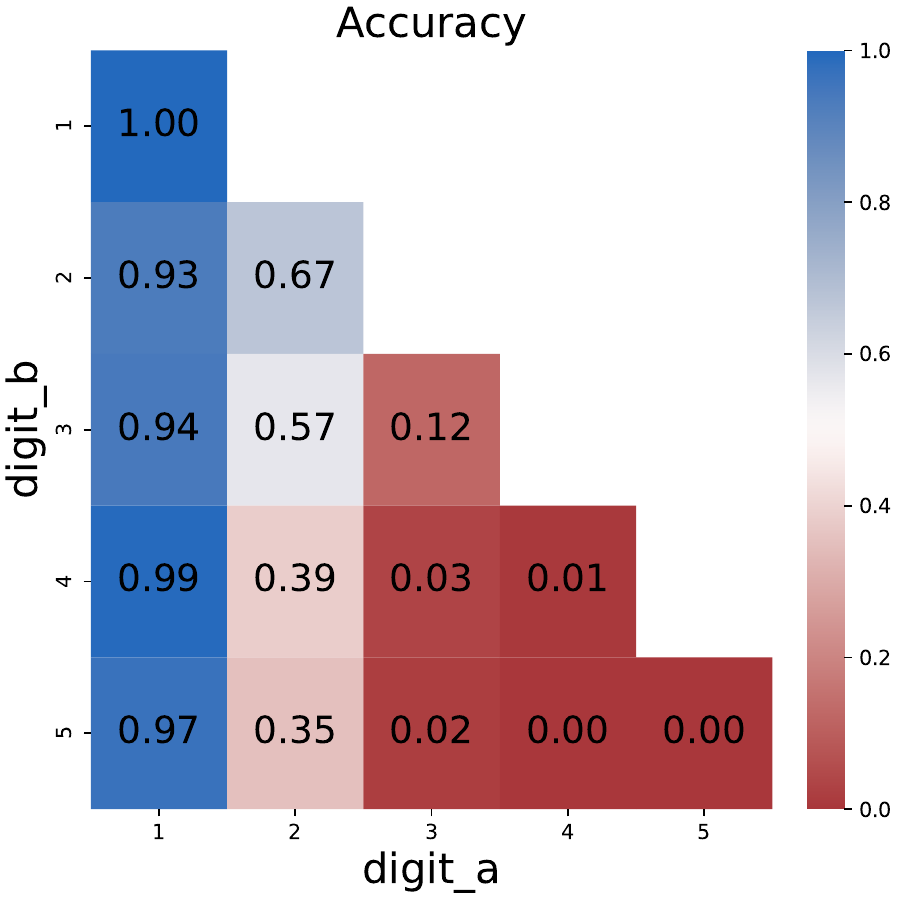}
        \caption{Plain multiplication}
        \label{subfig:multi_main_plain}
    \end{subfigure}
    \hfill
    \begin{subfigure}[b]{0.23\textwidth}
        \includegraphics[width=\linewidth]{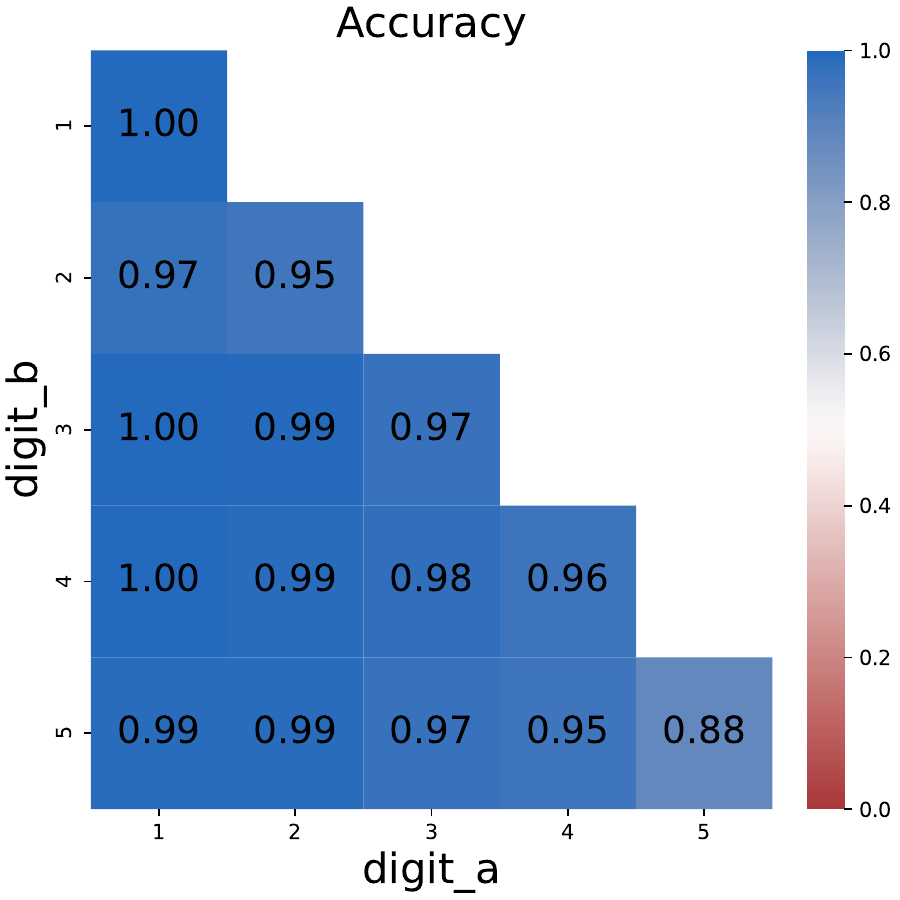}
        \caption{CoT multiplication}
        \label{subfig:multi_main_cot}
    \end{subfigure}
    \hfill
    \begin{subfigure}[b]{0.23\textwidth}
        \includegraphics[width=\linewidth]{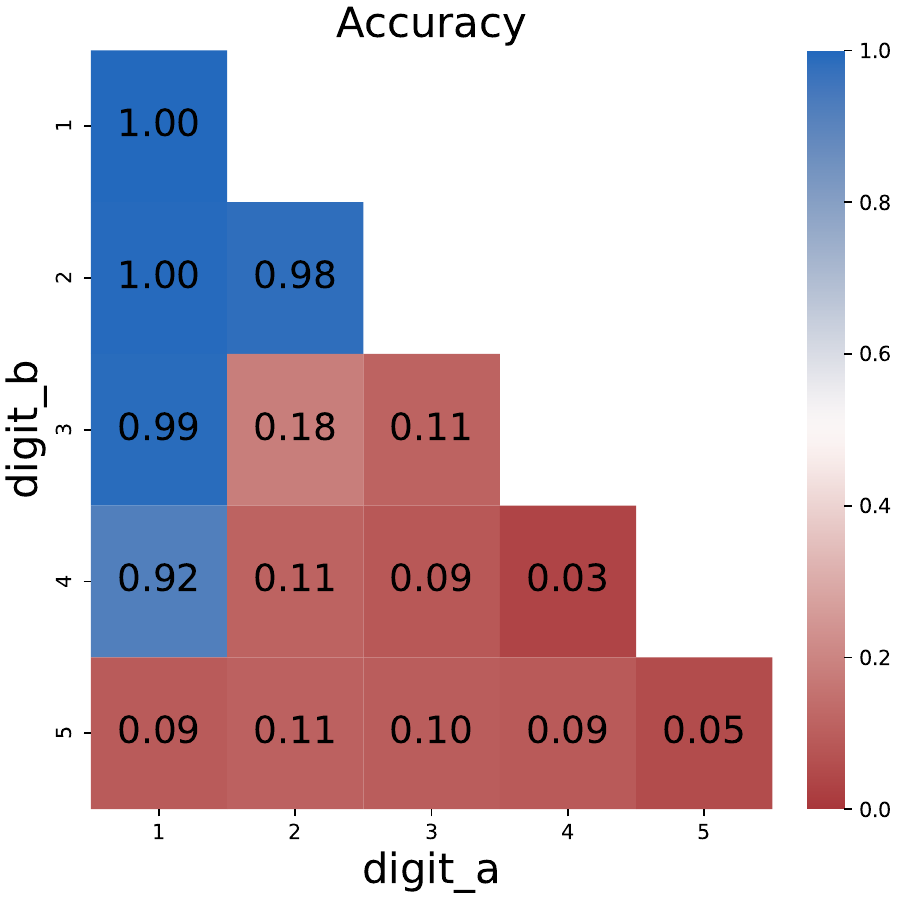}
        \caption{Plain DP}
        \label{subfig:dp_main_plain}
    \end{subfigure}
    \hfill
    \begin{subfigure}[b]{0.23\textwidth}
        \includegraphics[width=\linewidth]{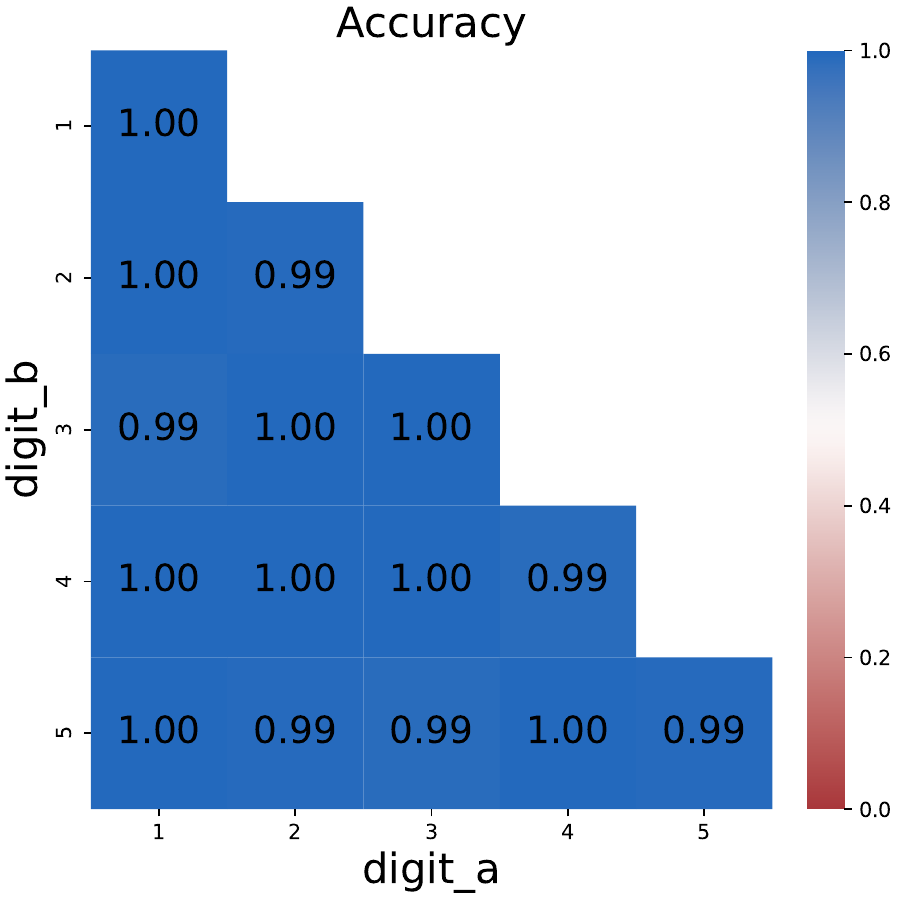}
        \caption{CoT DP}
        \label{subfig:dp_main_cot}
    \end{subfigure}
    \caption{Comparison on model accuracy between plain prompting and chain-of-thought prompting.}
    \label{fig:main_results}
\end{figure*}

\section{CoT Tokens Store Intermediate Results}
\label{sec:cot_are_results}
\paragraph{Experimental Setup}
\label{ssec:setup}
In all of our experiments, We use Qwen-2.5-1.5B~\cite{yang2024qwen2} as the backbone model.
On each task, we finetune the model on the corresponding training data and then evaluate whether the generated final answer matches the golden answer.
The training stage goes under a learning rate of 1e-5 for 1 epoch.
See Appendix \ref{sec:appendix_main} for detailed hyperparameter settings.

\subsection{Necessity of Chain-of-Thoughts}
\label{ssec:cot_necessity}
We start by examining the effectiveness of CoT by comparing the model performance under direct prompting and CoT settings.
As illustrated in Figure \ref{fig:main_results}, training the model with direct prompts faces difficulty starting from 3*3 multiplication problems, and completely fails on larger numbers.
In contrast, the model could easily solve multiplication problems with chain-of-thoughts, with near-perfect accuracy.

The same applies to dynamic programming problems.
Direct prompting would fail as the number of intermediate states increases, while CoT maintains its competence.
These results support the conclusion that chain-of-thoughts is necessary for solving inherent serial problems that require multi-step reasoning, just as previous research suggests~\cite{lichain, chen2024theoretical}.

\begin{figure}[htbp]
    \centering
    \includegraphics[width=0.45\linewidth]{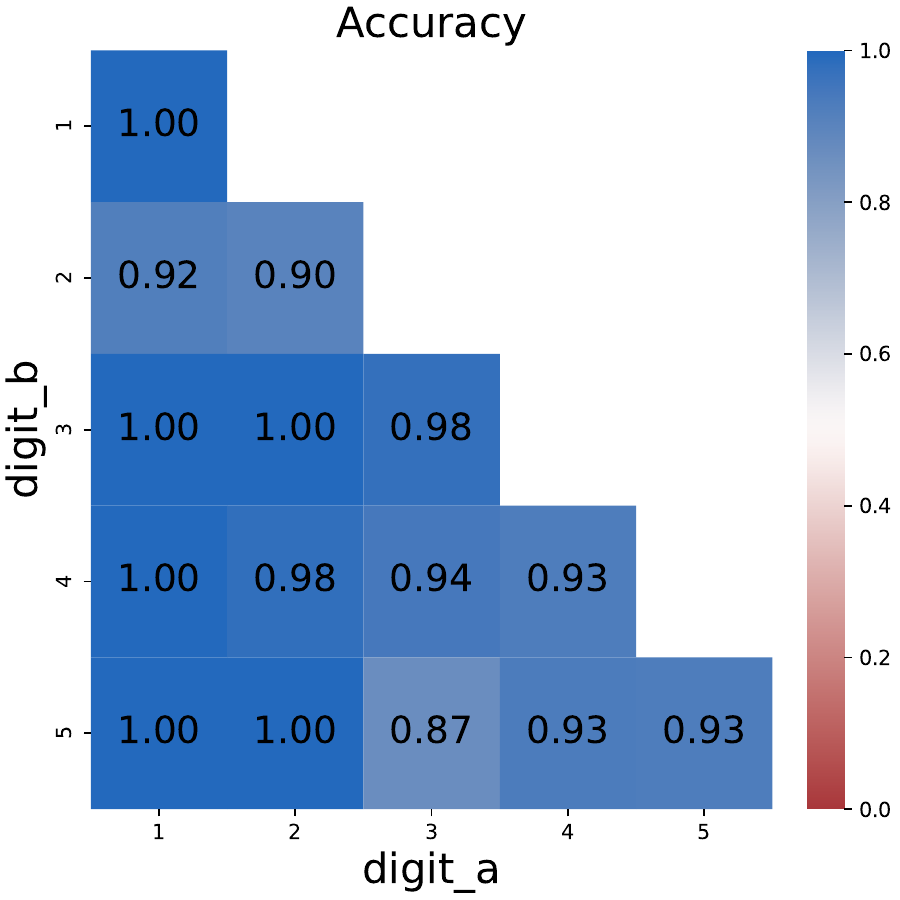}
    \label{subfig:multi_compressed}
    \caption{Model performance when non-result tokens are removed from CoT in multi-digit multiplication. Removing these tokens has little impact.}
    \label{fig:compressed_results}
\end{figure}

% \subsection{Redundancy of Non-result Tokens} % TODO: 弱化一下
\subsection{Removing Non-result Tokens}
\label{ssec:result_tokens}
One of the concerns about CoT is whether it could be compressed into a more concise form.
An obvious approach is to remove some less important tokens.
To be specific, we remove tokens that are neither a number nor a symbol\footnote{For example word ''carry`` in multiplication problems, see Appendix \ref{sec:appendix_main} for details}, making CoT a sequence consisting purely of intermediate results to solve the task.

Figure \ref{fig:compressed_results} shows the model performances after removing these tokens.
While the removal would make the CoT unreadable, models finetuned on compressed CoT still achieve satisfying performance, even outperforming the full CoT setting.
We can infer from this phenomenon that intermediate results are more important than semantic completeness in CoT.
In other words, the central function of CoT is to store the sequence of intermediate results.

\begin{figure*}[htbp]
    \centering
    \includegraphics[width=0.90\textwidth]{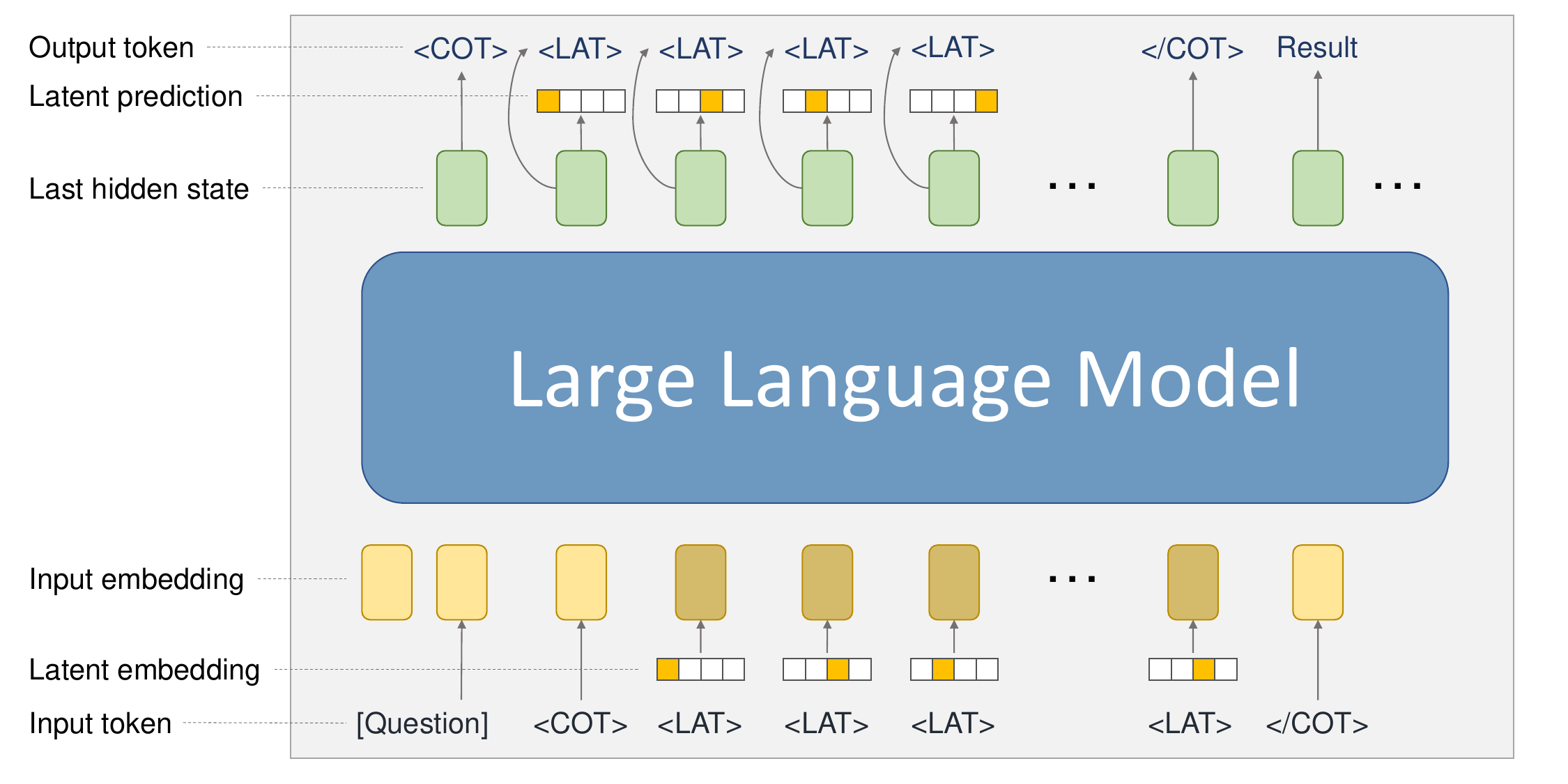}
    \caption{The model structure used to reason with latent tokens. We use one-hot vectors as the latent embedding of latent tokens \texttt{<LAT>}. When the input token is a latent token, we use its projected latent embedding to replace the original input embedding. Correspondingly, a latent output head is added to predict the latent embedding of the next token from the last hidden state.}
    \label{fig:latent_datapath}
\end{figure*}

\begin{figure}[htbp]
    \centering
    \begin{subfigure}[b]{0.22\textwidth}
        \includegraphics[width=\linewidth]{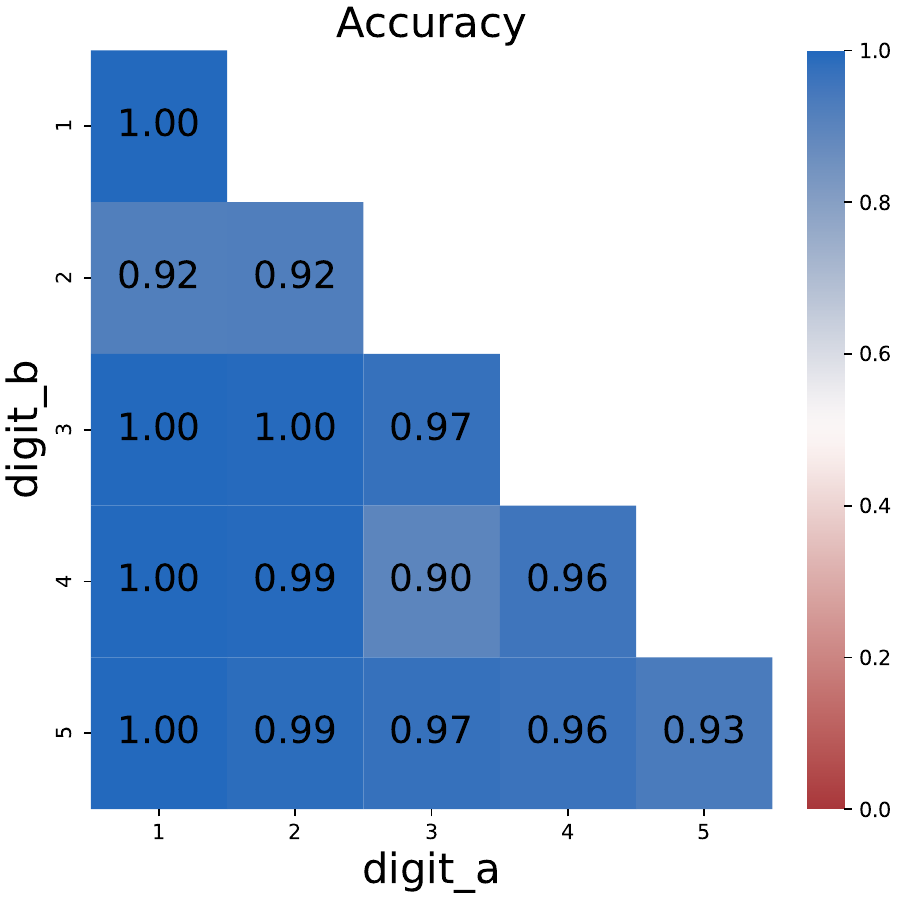}
        \caption{Multiplication}
        \label{subfig:multi_latent}
    \end{subfigure}
    \hfill
    \begin{subfigure}[b]{0.22\textwidth}
        \includegraphics[width=\linewidth]{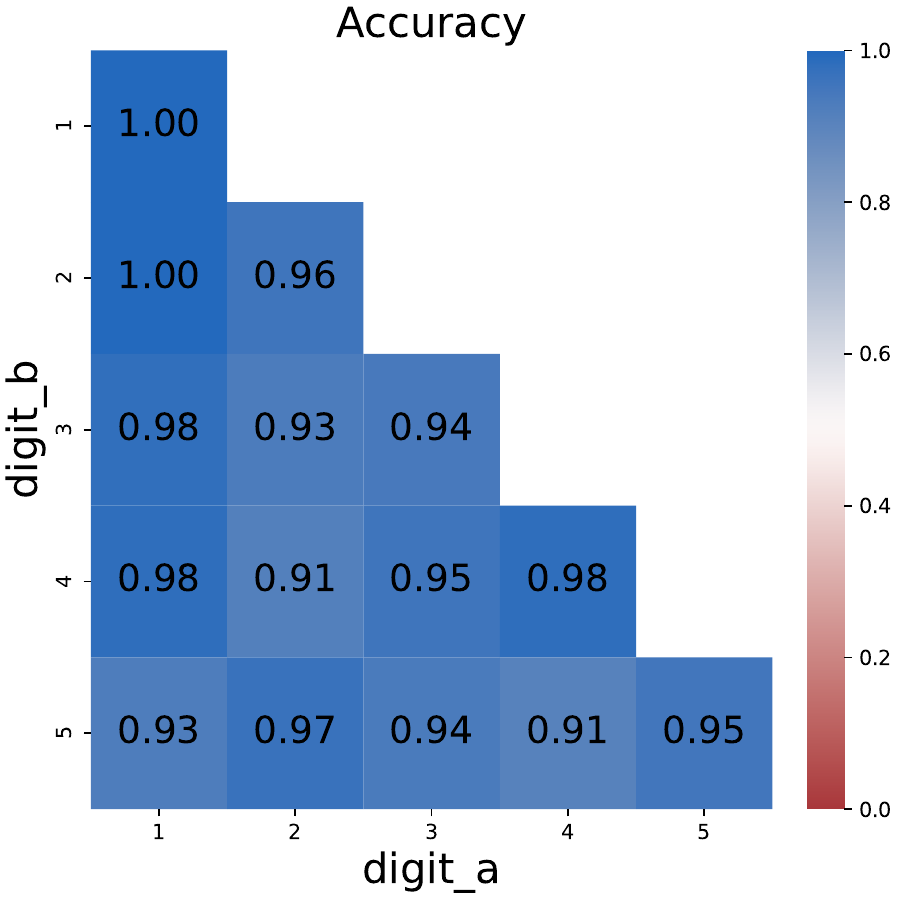}
        \caption{DP}
        \label{subfig:dp_latent}
    \end{subfigure}
    \caption{Model performances when merging intermediate results into latent tokens.}
    \label{fig:latent_results}
\end{figure}

% \paragraph{Merging intermediate results into latent tokens.}
\subsection{Merging Results into Latent Tokens} % TODO: latent目前看起来有种变种LM的样子需要修改
\label{ssec:latent_tokens}
Another concern about CoT is whether intermediate results should be explicitly recorded.
To test this hypothesis, we try to merge some of the intermediate results, and represent the merged results with latent tokens.

\paragraph{Method design.}
As depicted in Figure \ref{fig:latent_datapath}, we use latent tokens \texttt{<LAT>} to store intermediate results, and each latent token stores the information of a complete number.
For simplicity, we use one-hot vectors $\mathbf{l}$ as the embedding of latent tokens:
a one-hot vector $\mathbf{l} = (l_1, l_2, \ldots, l_d) \in \mathbb{R}^d$ consisting of $d$ dimensions could represent a number $N$ of at most $n$ digits, where $d = 10n$.
\begin{equation}
    l_{10k+x} = 
    \begin{cases}
    1, \lfloor \frac{N}{10^{k}} \rfloor \mod 10 = x \\
    0, \lfloor \frac{N}{10^{k}} \rfloor \mod 10 \neq x
    \end{cases}
\end{equation}
We start by setting all values in $L$ to 0.
Assuming that the value of the $k$-th digit under the little-endian system is $x$, we set $l_{10k+x} = 1$.
In this way, we could represent a number with a single latent token instead of multiple tokens.

To support reasoning with latent tokens, we augment the Transformer structure by adding an input projection module $P_{in}$ and a latent output head $P_{out}$.
When the input token $c_t$ at position $t$ is a latent token, we feed its latent embedding $\mathbf{l}_t$ to the projection module $P_{in}$, and use the projected vector as the input embedding;
Correspondingly, the last hidden state $\mathbf{h}_t$ is fed to the latent output head $P_{out}$ aside from the default LM head to predict the latent embedding of the next token $\mathbf{l}_{t+1}$.

We use linear layers to implement $P_{in}$ and $P_{out}$, which can be described as:
\begin{align}
    P_{in}(\mathbf{l}_{t}) &= \mathbf{W}_{in}\mathbf{l}_{t} + \mathbf{b}_{in} \\
    P_{out}(\mathbf{h}_{t}) &= \mathbf{W}_{out}\mathbf{h}_{t} + \mathbf{b}_{out}
\end{align}
where $\mathbf{W}_{in}, \mathbf{b}_{in}, \mathbf{W}_{out}, \mathbf{b}_{out}$ are trainable parameters.
We randomly initialize these parameters.

% An additional latent loss $\mathcal{L}_{lat}$ is introduced to train these modules:
An additional latent loss $\mathcal{L}_{lat}$ is introduced to train the augmented model:
\begin{equation}
    \mathcal{L}_{lat} = \frac{1}{N_{l}} \sum_{c_t = <LAT>} \text{BCE}(\sigma(P_{out}(\mathbf{h}_{t}), \mathbf{y})) 
\end{equation}
Where $N_{l}$ is the number of latent tokens, $\mathbf{y}$ is the golden latent embedding, BCE is the binary cross entropy loss function, and $\sigma$ is the Sigmoid function.

\begin{figure*}[htbp]
    \centering
    \begin{subfigure}[b]{0.46\textwidth}
        \includegraphics[width=\linewidth]{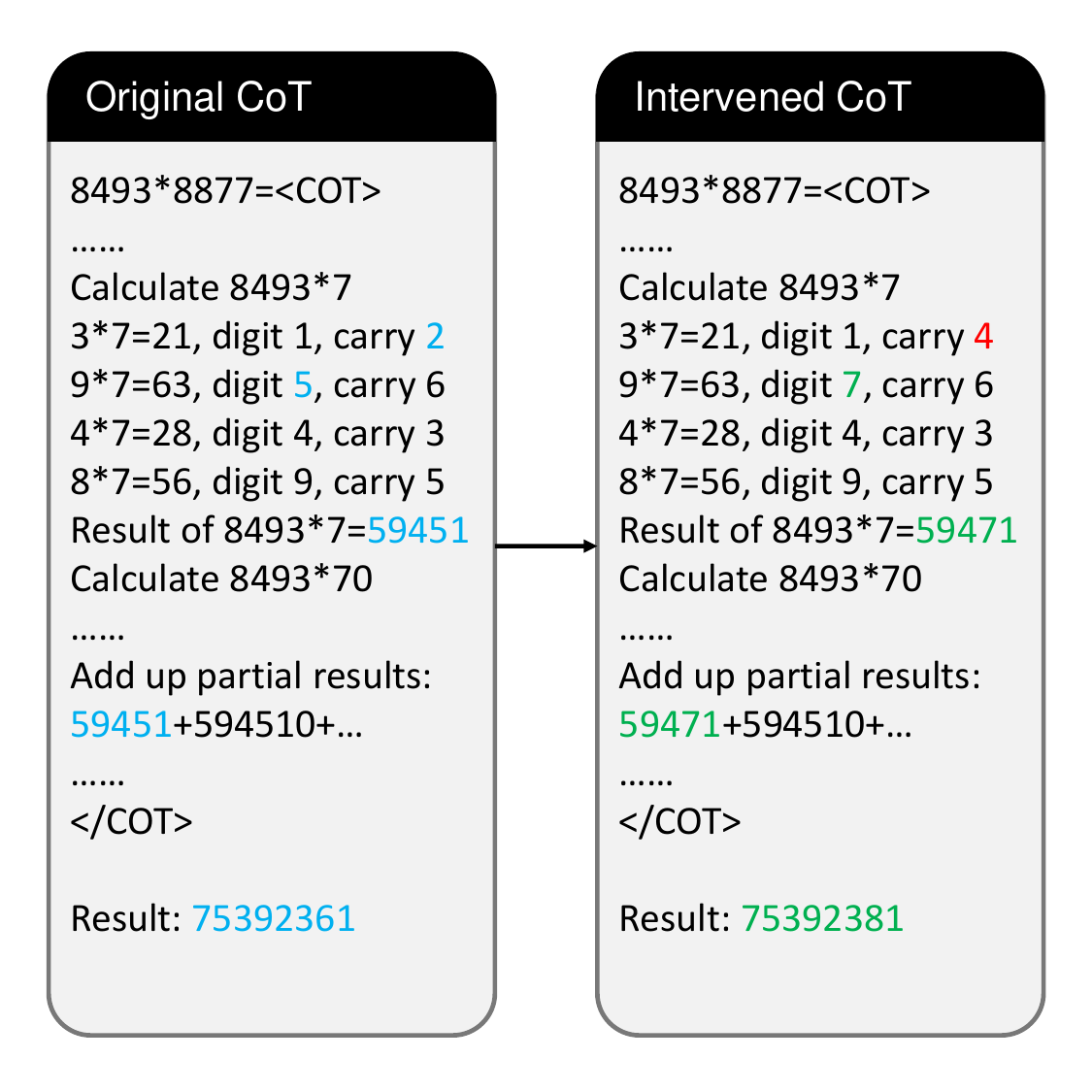}
        \caption{Successful intervention}
        \label{subfig:successful_intervention}
    \end{subfigure}
    \hfill
    \begin{subfigure}[b]{0.46\textwidth}
        \includegraphics[width=\linewidth]{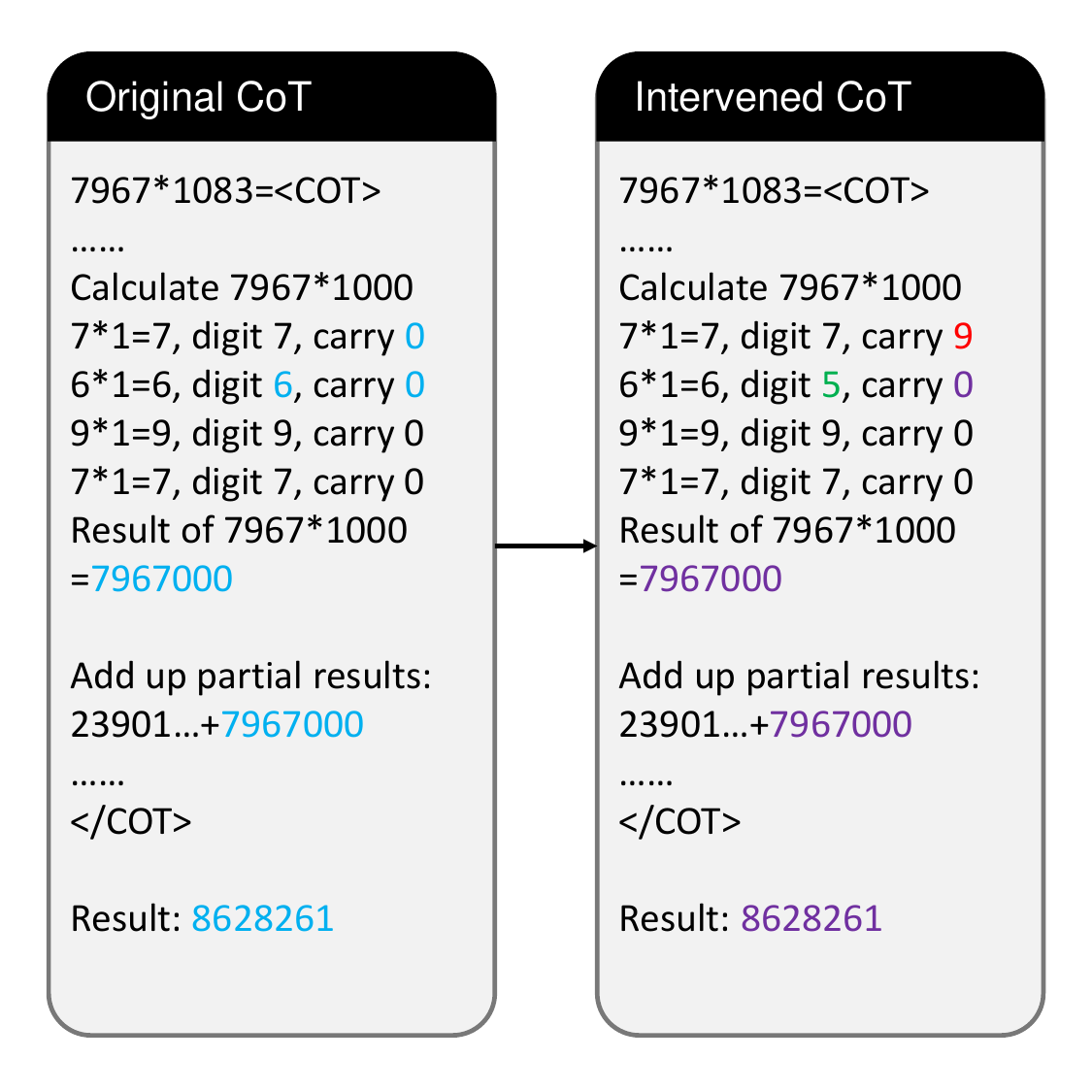}
        \caption{Intervention with a shortcut error}
        \label{subfig:shortcut_intervention}
    \end{subfigure}
    \caption{Examples of a successful intervention (left) and an intervention with a shortcut error (right). \textcolor{Blue}{Blue} numbers refer to relevant values in the original CoT, \textcolor{Red}{red} numbers refer to the intervention, \textcolor{Green}{green} numbers refer to values that change as expected, but \textcolor{Purple}{purple} numbers do not change due to a shortcut error.}
    \label{fig:intervention_examples}
\end{figure*}

\paragraph{Experimental setup.}
For multiplication problems, we replace each digit-wise multiplication step with a single latent token and set $d = 20$;
For DP problems, we replace each intermediate state with a single latent token and set $d = 50$.
We add the latent loss $\mathcal{L}_{lat}$ with the default LM head loss as the final loss for training.

Figure \ref{fig:latent_results} shows the model performances when trained with latent tokens.
Surprisingly, merging digit tokens to a single latent token does not detriment the performance: the model retains most of its ability to solve problems.
The accuracy of using latent tokens on multiplication problems is almost identical with the accuracy of using full CoT.
On 5*5 multiplication problems, using latent tokens even surpasses the original CoT, suggesting that the form of intermediate results does not matter.

However, it can also be observed that using latent tokens brings disadvantage on DP problems where latent tokens store larger numbers.
For example, the accuracy reduces by 9\% on 4*5 DP problems.
This raises the hypothesis that the computation complexity should not exceed a certain limit, which we will discuss further in Section \ref{ssec:probing}.

\section{CoT Tokens are Mutable Variables}
\label{sec:cot_are_variables}
In the previous section, we find that while CoT is essential for solving complex problems (Section \ref{ssec:cot_necessity}), the tokens representing intermediate results are more important than others (Section \ref{ssec:result_tokens}).
Meanwhile, compressing intermediate results into latent tokens would not obviously harm model performance (Section \ref{ssec:latent_tokens}), indicating that intermediate results could be stored in different forms.

Here, we continue to discuss whether these stored intermediate results are causally related to the final prediction, and how the computation complexity between intermediate results affects model performance.

\begin{figure}[htbp]
    \centering
    \begin{subfigure}[b]{0.22\textwidth}
        \includegraphics[width=\linewidth]{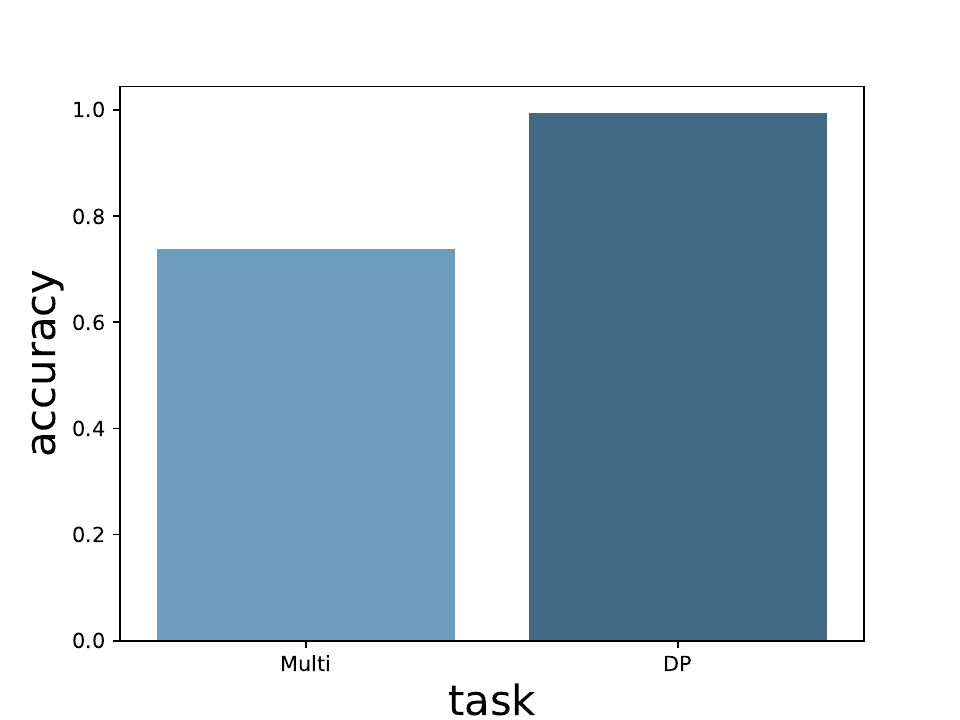}
        \caption{Success rate}
        \label{subfig:intervention_success_rate}
    \end{subfigure}
    \hfill
    \begin{subfigure}[b]{0.22\textwidth}
        \includegraphics[width=\linewidth]{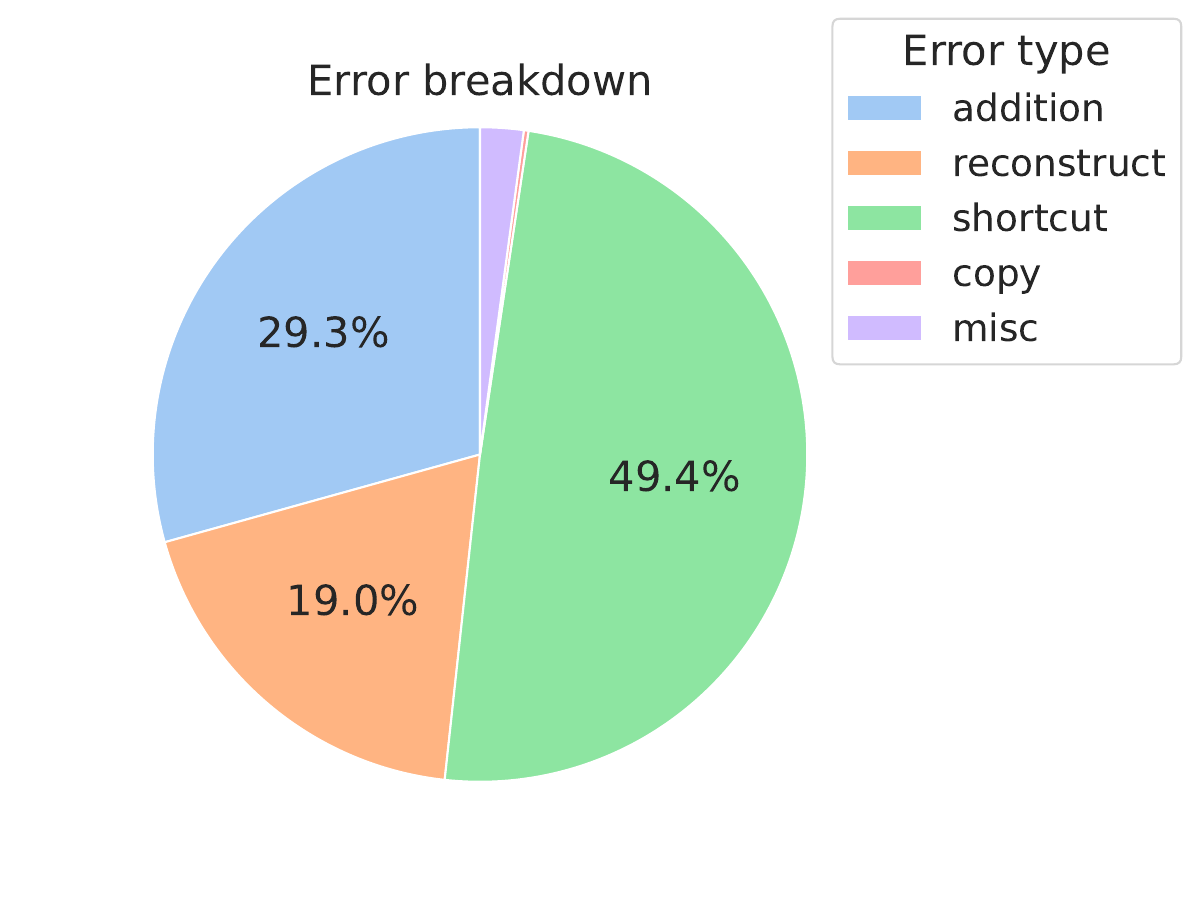}
        \caption{Error breakdown}
        \label{subfig:intervention_breakdown}
    \end{subfigure}
    \caption{(a) \textbf{Success rate of intervention}. When the intervened output is the same as simulated, we view it as a successful intervention. (b) \textbf{Error breakdown}. Shortcut error occupies a large percentage of the errors.}
    \label{fig:intervention_results}
\end{figure}

\begin{figure}[htbp]
    \centering
    \includegraphics[width=0.45\textwidth]{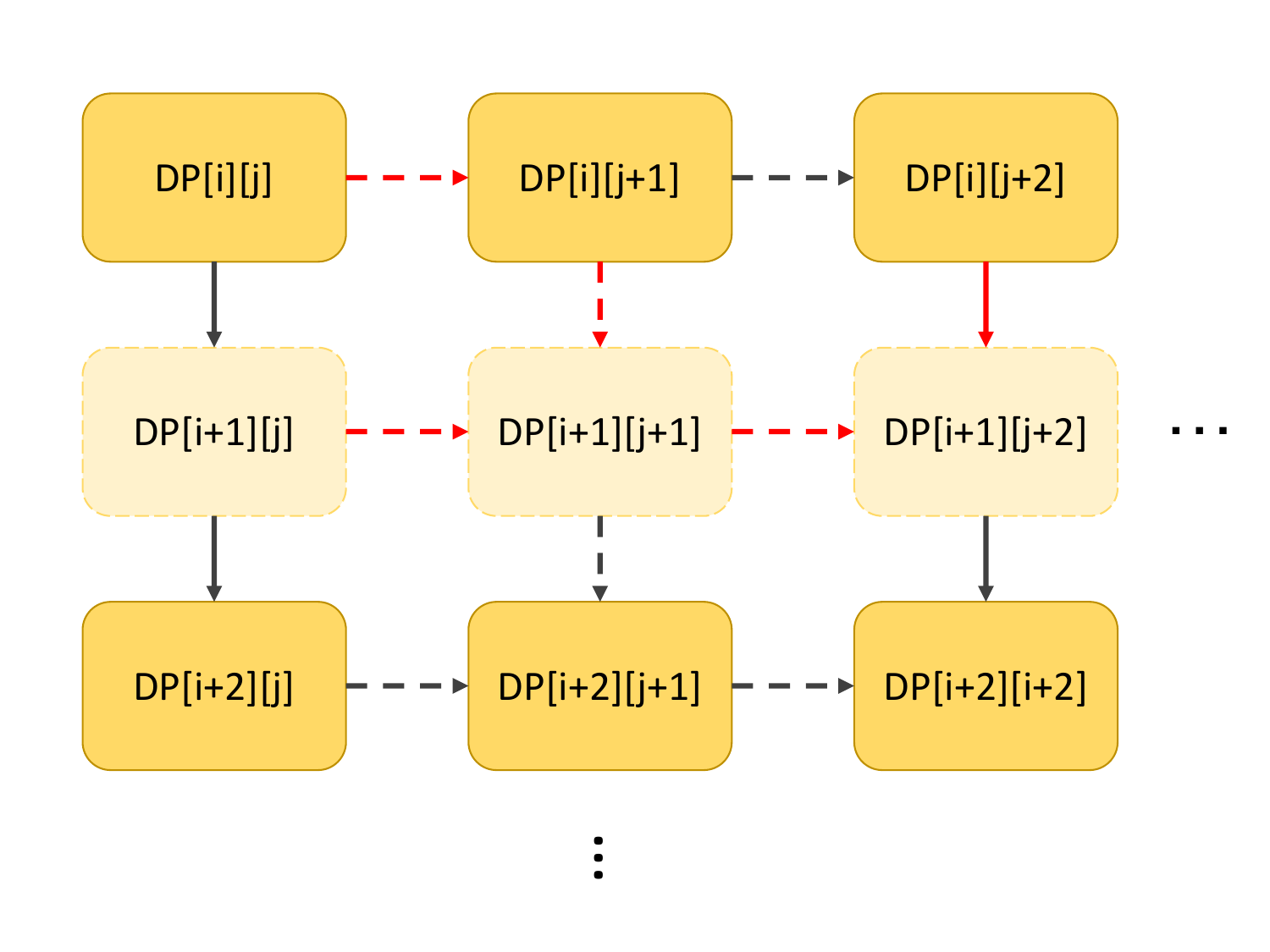}
    \caption{Demonstration of the alternative merging strategy. Each line refers to the compare-then-add state transfer function in the original setup. Nodes corresponding to the dashed boxes will not appear in the new CoT. It will cost at most 3 compare-then-add operations (\textcolor{Red}{red} lines) to transfer states between new matrix tokens.}
    \label{fig:probing_demo}
\end{figure}

\subsection{Intervening the Value of CoT Tokens}
\label{ssec:intervention}
An inherent problem is that while CoT is essential for reaching the right answer, some of the intermediate results may only be correlational to the output, rather than having causal effects.
To address this problem, we perform intervention experiments by replacing intermediate results and observe whether the final result would change as expected.

For multiplication problems, we randomly choose a substep in CoT and replace its result with a different random number;
For DP problems, we randomly choose an intermediate state and replace it with a different random number.
For simplicity, we perform interventions on 4*4 problems, and only one number is replaced in each data entry.
Details are described in Appendix \ref{sec:appendix_intervention}.

As shown in Figure \ref{subfig:intervention_success_rate}, the intervention on both tasks achieves a decent success rate, clearly indicating that the intermediate values stored in CoT tokens are causally related to the final answer.
We also notice that subsequent reasoning steps will change correspondingly. 
Take Figure \ref{subfig:successful_intervention} as an example, when we change the carry from 2 to 4, the model generates a result of $8493*7=59471$ instead of 59451, just as simulated.
In other words, tokens in CoT not only store intermediate values, but they are also ``variables'' that would affect subsequent reasoning steps.

Another interesting observation is that the success rate on multiplication problems is significantly lower than that on DP problems.
We investigate the cause of unsuccessful interventions and categorize them into 5 categories.
(1) \textbf{Addition error} means that the model fails to add up partial multiplication results;
(2) \textbf{Reconstruction error} means that the partial multiplication result conflicts with digit-wise results;
(3) \textbf{Copy error} means that partial multiplication results do not correctly appear in the addition step; 
(4) \textbf{Shortcut error} means that the model learns a ``shortcut'' on certain multiplications (usually when one of the operands is 0 or 1);
(5) \textbf{Misc error} covers the remaining errors.

Figure \ref{subfig:intervention_breakdown} illustrates the distribution of error types.
Among the 5 types, shortcut error occupies the largest portion.
As shown in Figure \ref{subfig:shortcut_intervention}, while changing the carry from 0 to 9 will affect the next digit as intended, the model does not change its result in the substep $7967*1000$.
When multiplying a number $x$ by 1, the model seems to be taking a shortcut of directly copying $x$, rather than collecting the digit-wise multiplication results.
% Intermediate results on these shortcuts are only ``printed'' but not used in following computations, leaving them like unused variables in computer programs.

To sum up, language models use the value in CoT tokens like treating program variables, but models may develop shortcut on easy subproblems that leave certain variables unused.

\begin{figure}[htbp]
    \centering
    \includegraphics[width=0.45\textwidth]{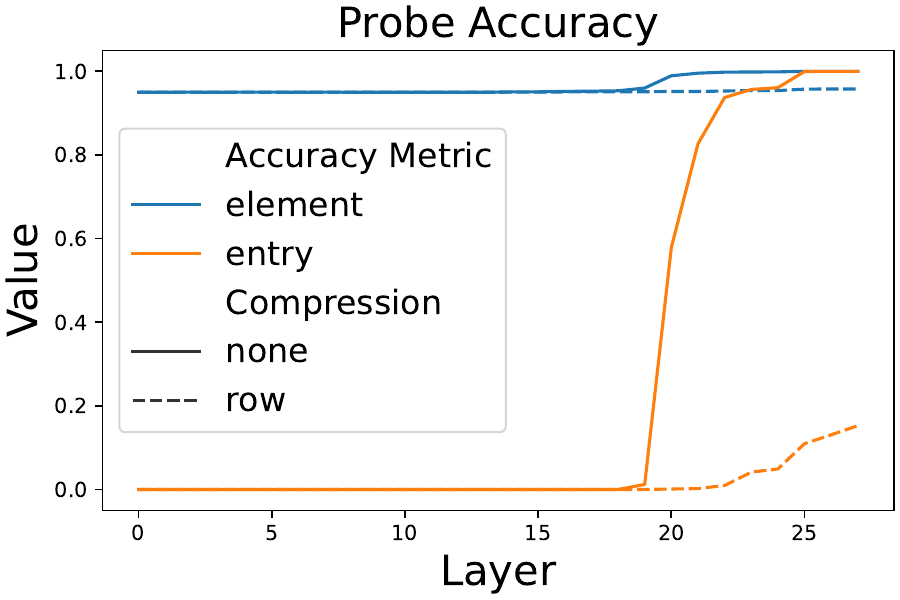}
    \caption{Probing accuracy on different layers. Intermediate variable values can only be probed on late layers, regardless of the overall accuracy.}
    \label{fig:probing_accuracy}
\end{figure}

\subsection{Probing the Limit of CoT Tokens}
\label{ssec:probing}
In Section \ref{ssec:latent_tokens}, we discover that intermediate values can be compressed in latent tokens.
This naturally raises the question: to what extent could the values be compressed?
To address this problem, we adopt some aggressive compression strategies and use linear probing classifiers to observe how the compression affects the final output.

We choose 5*5 DP problems as the base problem and use the latent token setting in Section \ref{ssec:latent_tokens}.
Specifically, we introduce an alternative strategy that merges two adjacent latent tokens in a row to one latent token (Figure \ref{fig:probing_demo}).
In this way, this strategy yields a 3*3 CoT token matrix instead of a 5*5 matrix.
However, the computational complexity between CoT tokens also increases: it would cost up to 3 times as much as in the original case.

For each CoT token \texttt{<LAT>}, we use a linear probe $P$ to probe its latent embedding $\mathbf{l}$ from the hidden states $\mathbf{h}_k$ on different layer $k$ of the previous token.
We use a unique probe $P_k$ for each layer:
\begin{equation}
    P_k(\mathbf{h}_k) = \mathbf{W}_k \mathbf{h}_k + \mathbf{b}_k
\end{equation}
where $\mathbf{W}_k$ and $\mathbf{b}_k$ are trainable parameters.

After training the probes on the training set, we evaluate them with two metrics:
\textbf{element accuracy} evaluates the ratio of correctly predicted individual dimensions, and \textbf{token accuracy} evaluates the ratio of correct latent tokens.

Figure \ref{fig:probing_accuracy} shows the result of probing CoT tokens.
Aggressively merging CoT tokens will significantly lower both element accuracy and token accuracy, meaning that there exists a computation complexity limit, over which the LLM can no longer correctly calculate the next intermediate variable.

\begin{figure}[htbp]
    \centering
    \includegraphics[width=0.45\textwidth]{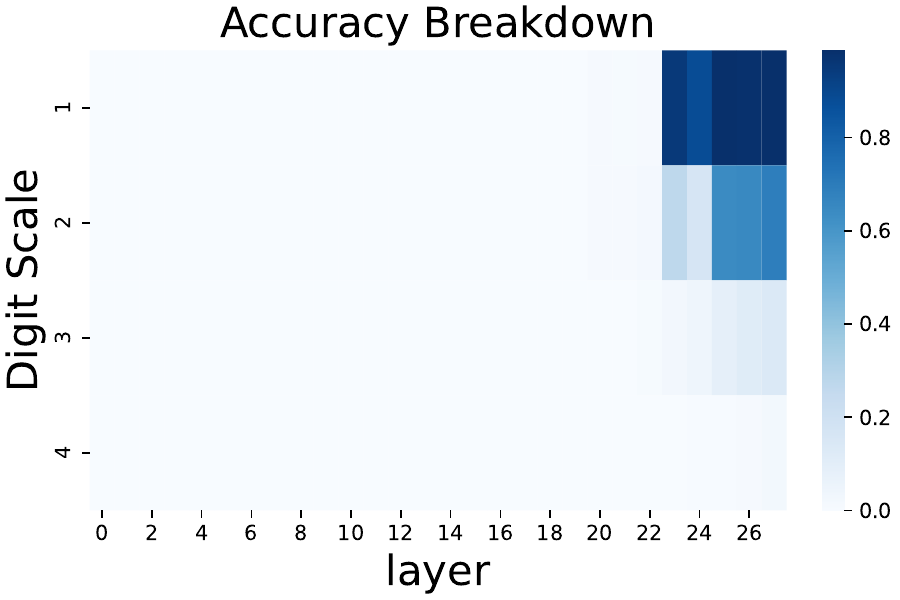}
    \caption{Accuracy breakdown by the scale of target values. When computational complexity between tokens exceeds a limit, the model will fail.}
    \label{fig:probing_breakdown}
\end{figure}

Figure \ref{fig:probing_breakdown} further breaks down the accuracy distribution by the range of values stored in merged latent tokens.
We can see that merging latent tokens has little impact on numbers with a digit length of 1 or 2, but would decrease the accuracy to near 0 on larger number values.
This phenomenon can be explained as it is easier to calculate small numbers, and thus the model could ``afford'' the extra computational cost of merging latent tokens. 

Another interesting point to notice is that two accuracy curves share a similar pattern:
the token accuracy stays at 0 from early layers, and rapidly rises around layer 20.
Previous work~\cite{stolfo2023mechanistic, zhu2025language} has concluded that LLMs tend to use early-mid layers to gather and process information from previous tokens, and determine the output only in mid-late layers.
We may further assume that the role of layers will not change with computation complexity between CoT tokens.

\section{Discussion}
% TODO
\paragraph{Explaining alternative CoT forms.}
By viewing CoTs as programs, we can explain alternative CoT forms in a novel way.
For example, the success of internalizing CoT steps~\cite{deng2024explicit, hao2024training} could be viewed as compressing explicit step tokens into implicit latent tokens that cover all essential intermediate values.
And the validity of inserting meaningless filler tokens~\cite{goyalthink, pfaulet} comes from enabling LLMs to store intermediate results in the hidden states of filler tokens.
% The commonality of these approaches is that they reserve critical intermediate variables, regardless of their CoT forms.
% As long as tokens reserve critical intermediate values, CoT can be represented in different forms.
By reserving space for complex reasoning steps and compressing simple reasoning steps, we could design high-efficiency CoTs.

\paragraph{Generalization of CoT ``programs''.}
From the experiments in previous sections, we can see that CoT tokens store intermediate values, and their values are subsequently used like the way variables function in computer programs.
Theoretical proof has been made that Transformer with a recurrent module is Turing complete~\cite{perez2021attention}.
However, there is also evidence that LLMs may struggle to generalize on compositional problems~\cite{dziri2023faith}: models trained on easy problems would fail on more complex problems.
In real-world settings, the type and complexity of desired programs are unknown, and a general-purpose LLM needs to first determine the type of program to use, or in other words, generate a meta-program first.
% It would be beneficial to explore the generalization ability of LLMs on different types of program paradigms, like loop, search, etc.

\paragraph{Identification of intermediate variable tokens.}
% It is not surprising that the CoT generated by LLMs is redundant and could be shortened via simple strategies like explicitly specifying length constraints in prompts.
It is not surprising that the CoT generated by LLMs is partially redundant and could be shortened.
In Section \ref{ssec:result_tokens}, we find that preserving value tokens could retain most of the ability of language models.
While it is easy to judge whether a token stores intermediate results in multiplication and DP problems, it is harder to identify variable tokens on general tasks:
\citet{madaan2022text} finds that plain text helps LLMs elicit semantic commonsense knowledge, which may be infused into later CoT tokens.
% If the CoT of natural language problems could be converted to a more structured form, it could be easier to identify variable tokens and benefit CoT compression.
Developing an approach to identifying variable tokens would benefit CoT compression.

\paragraph{Estimation of computational complexity between variable tokens.}
Section \ref{ssec:probing} shows that LLMs would fail when the computational complexity between variable tokens exceeds a certain limit.
However, it is difficult to estimate the exact complexity limit for LLMs.
It is possible to calculate the theoretical bound of ability for finite-precision Transformers~\cite{chen2024theoretical}, but how LLMs process semantic information is still largely opaque, and unexpected features may appear~\cite{lindsey2025biology}.
Moreover, LLMs are not guaranteed to solve similar subproblems in the same way, they may take shortcuts (Section \ref{ssec:intervention}) that would largely affect the computational complexity between variable tokens.
We hope that the broader research community could help estimate the computational complexity between variable tokens in different types of questions.

\section{Related Work}
\subsection{Chain-of-Thought Reasoning}
Chain-of-Thoughts (CoT)~\cite{wei2022chain} is a commonly adopted technique in LLMs.
Nowadays, CoT refers to a broad range of approaches that require LLMs to generate an intermediate reasoning process before reaching the final answer.
Typical approaches include designing the prompt~\cite{wei2022chain, khotdecomposed, zhouleast} and finetuning LLMs on existing chain-of-thoughts~\cite{yuemammoth, yumetamath}.
Recently, reinforcement learning also reveals its great potential in enabling LLMs to perform complex reasoning without extensive human annotations~\cite{havrilla2024teaching, wang2024math, shao2024deepseekmath, guo2025deepseek}.
While the tokens in CoT can be classified into symbols, patterns, and text, which both contribute to the final answer~\cite{madaan2022text}, it seems that LLMs can still perform well with a small amount of CoT tokens~\cite{xu2025chain}.

Aside from plain text, researchers have also explored alternative forms of CoT.
Some works focus on search abilities, like tree-form thought traces~\cite{yao2023tree, xie2023self} and Monte-Carlo Tree Search (MCTS) algorithms~\cite{zhang2024llama, guan2025rstar}.
Another line of work attempts to reason in a latent space:
~\citet{goyalthink} uses a pause token to help models process extra computation before reaching an answer, and~\citet{pfaulet} shows it is also possible to replace CoT with meaningless filler tokens.
On top of this, ~\citet{deng2024explicit} tries to train models with gradually shortened CoT, and COCONUT~\cite{hao2024training} proposes the continuous thought paradigm, where the last hidden state of a latent token is used as the next input embedding.

\subsection{Theoretical Analysis on CoT}
It has been noticed that LLMs face difficulty in compositional problems where combining multiple reasoning steps is strictly required, and it may be an intrinsic drawback of the Transformer structure~\cite{dziri2023faith}.
\citet{feng2023towards} explains the phenomenon with the circuit complexity theory, and reaches the conclusion that it is impossible for a constant-depth log-precision transformer to solve certain math problems like linear equations.
However, with the help of CoT, the model could solve these problems in polynomial complexity.
\citet{lichain} further extends the conclusion that constant-depth transformers using constant-bit precision could solve any problems solvable by boolean circuits, as long as they are equipped with CoT whose steps are longer than the circuit size.
\citet{chen2024theoretical} analyzes the problem with a multi-party autoregressive communication model, and finds that it is exponentially harder for Transformer models to solve composition tasks that require more steps than the model layers, and CoT could make the problem exponentially easier.

In fact, Transformer models are powerful enough to represent finite-state automata~\cite{liutransformers}, and could even be Turing-complete~\cite{perez2021attention} to simulate computer programs when equipped with loop modules~\cite{giannou2023looped}.
We hold the belief that these findings could also be extended to chain-of-thoughts reasoning.

\section{Conclusion}
In this paper, we empirically explore the role CoT tokens play in reasoning.
By observing the model performance on multi-digit multiplication problems and dynamic programming, we confirm that CoT is essential for solving these compositional problems.
We further find that we could mostly preserve model ability by only using tokens that store intermediate results, and these intermediate results could be stored in different forms like latent token embeddings.

To validate the causal connection between CoT tokens and model output, we randomly replace some values in CoT, and find that both the subsequent reasoning process and the final result would change corresponding to the intervention.
The way CoT tokens behave is similar to the function of computer program variables. 
However, in easy subproblems LLMs would learn shortcuts that are unfaithful to the generated reasoning process, and the intervention would fail under these scenarios.
We also train probing classifiers to probe variable values from hidden states on different layers, and find that there exists a computational complexity limit between CoT tokens.
Intermediate values could be compressed within a single latent CoT token, but the model would drastically fail when computational complexity exceeds the limit.

Our work conducts preliminary experiments on the function of CoT tokens, and there still exist mysteries like generalization ability, variable identification and complexity limit estimation, which we leave for future explorations.

\section*{Limitations}
In this paper we empirically demonstrate that an important function of CoT tokens is to store intermediate values, and these values function like program variables.
However, currently we are not able to give a theoretical proof on this statement.

Another limitation of our work is that the experiments are conducted on two synthetic tasks with Qwen-2.5-1.5B, as it is difficult to identify and analyze intermediate results in real-world datasets like GSM8K and Math.
Future experiments on other problems and models will be beneficial.

\bibliography{custom}

\appendix

\section{Details on Multiplication Task}
\label{sec:appendix_multi}
\paragraph{Dataset construction}
For each problem scale of $m \times n$ that multiplies $m$-digit number $a$ with $n$-digit number $b$, we generate 100,000 data entries by randomly sampling $a$ and $b$.
When the scale is small (for example $1 \times 2$), we exhaustively generate all number pairs instead.
The generated data entries are then divided into train and test splits with a ratio of 90\%/10\%.

\paragraph{Prompt and CoT Formulation}
We use simple multiplication expressions as prompts.
Figure \ref{fig:app_multi_prompt} shows an example prompt for querying the model to perform multiplication.

For convenience, we use \texttt{<tool\_call>} as the start-of-CoT token \texttt{<COT>}, \texttt{</tool\_call>} as the end-of-CoT token \texttt{</COT>}, and \texttt{<|fim\_middle|>} as the latent token \texttt{<LAT>}, which already exist in the tokenizer vocabulary.

We formulate the reasoning process with the algorithm of digit-wise multiplication, whose example is demonstrated in Figure \ref{fig:app_multi_cot}.
In the compressed CoT setting, we remove all tokens that merely represent text semantics in CoT, namely ``Calculate'', ``digit'', ``carry'', ``Result of'' ``Add up partial results:'' and ``The final result is:'', whose example is demonstrated in Figure \ref{fig:app_multi_compressed}.

\begin{figure}[htbp]
    \centering
    \begin{tcolorbox}[
        title=Prompt example,
        coltitle=white,
        colbacktitle=black,
        colback=gray!10,
        colframe=gray!50,
        width=0.48\textwidth,
    ]
    3773*6821=
    \end{tcolorbox}
    \caption{Example prompt for the multi-digit multiplication task.}
    \label{fig:app_multi_prompt}
\end{figure}

\begin{figure*}[htbp]
    \centering
    \begin{tcolorbox}[
        title=Full CoT example,
        coltitle=white,
        colbacktitle=black,
        colback=gray!10,
        colframe=gray!50,
        width=0.95\textwidth,
    ]
3773*6821=<tool\_call>Calculate 3773*1 \\
3*1=3, digit 3, carry 0 \\
7*1=7, digit 7, carry 0 \\
7*1=7, digit 7, carry 0 \\
3*1=3, digit 3, carry 0 \\
Result of 3773*1=3773 \\
Calculate 3773*20 \\
3*2=6, digit 6, carry 0 \\
7*2=14, digit 4, carry 1 \\
7*2=14, digit 5, carry 1 \\
3*2=6, digit 7, carry 0 \\
Result of 3773*20=75460 \\
Calculate 3773*800 \\
3*8=24, digit 4, carry 2 \\
7*8=56, digit 8, carry 5 \\
7*8=56, digit 1, carry 6 \\
3*8=24, digit 0, carry 3 \\
Result of 3773*800=3018400 \\
Calculate 3773*6000 \\
3*6=18, digit 8, carry 1 \\
7*6=42, digit 3, carry 4 \\
7*6=42, digit 6, carry 4 \\
3*6=18, digit 2, carry 2 \\
Result of 3773*6000=22638000 \\
\\
Add up partial results: 3773+75460+3018400+22638000 \\
3773+75460+3018400+22638000=79233+3018400+22638000 \\
79233+3018400+22638000=3097633+22638000 \\
3097633+22638000=25735633 \\
\\
The final result is: 3773*6821=25735633</tool\_call> \\
\\
Result: 25735633
    \end{tcolorbox}
    \caption{Example CoT for the multi-digit multiplication task.}
    \label{fig:app_multi_cot}
\end{figure*}

\begin{figure*}[htbp]
    \centering
    \begin{tcolorbox}[
        title=Compressed CoT example,
        coltitle=white,
        colbacktitle=black,
        colback=gray!10,
        colframe=gray!50,
        width=0.95\textwidth,
    ]
3773*6821=<tool\_call>3773*1 \\
3*1 3 0 \\
7*1 7 0 \\
7*1 7 0 \\
3*1 3 0 \\
3773*1=3773 \\
3773*20 \\
3*2 6 0 \\
7*2 4 1 \\
7*2 5 1 \\
3*2 7 0 \\
3773*20=75460 \\
3773*800 \\
3*8 4 2 \\
7*8 8 5 \\
7*8 1 6 \\
3*8 0 3 \\
3773*800=3018400 \\
3773*6000 \\
3*6 8 1 \\
7*6 3 4 \\
7*6 6 4 \\
3*6 2 2 \\
3773*6000=22638000 \\
\\
3773+75460+3018400+22638000 \\
3773+75460+3018400+22638000=79233+3018400+22638000 \\
79233+3018400+22638000=3097633+22638000 \\
3097633+22638000=25735633 \\
\\
3773*6821=25735633</tool\_call> \\
\\
Result: 25735633
    \end{tcolorbox}
    \caption{Example CoT after compression for the multi-digit multiplication task.}
    \label{fig:app_multi_compressed}
\end{figure*}

\section{Details on Dynamic Programming Task}
\label{sec:appendix_dp}
\paragraph{Dataset construction}
Similar to the multiplication problems, we generate 100,000 data entries for each problem scale of $m \times n$ (whose input matrix has a shape of $m$ rows, $n$ columns), and divide them into train and test splits with a ratio of 90\%/10\%.

To control the value of intermediate states within a reasonable range, we ensure all values $x$ in the input matrix satisfy $1 < x < 100$.
In other words, each input value is a 2-digit number.

\paragraph{Prompt formulation}
We use a matrix whose shape is the same as the input matrix to store intermediate values.
The choice of special tokens \texttt{<COT>}, \texttt{</COT>} and \texttt{<LAT>} are the same as those in multiplication problems.

An example of the input prompt is shown in Figure \ref{fig:app_dp_prompt}, and an example of the full prompt is shown in Figure \ref{fig:app_dp_cot}.
Notice that we do not have a compressed version of CoT in dynamic programming tasks.

\begin{figure*}[htbp]
    \centering
    \begin{tcolorbox}[
        title=Prompt example,
        coltitle=white,
        colbacktitle=black,
        colback=gray!10,
        colframe=gray!50,
        width=0.95\textwidth,
    ]
Find a path in the given table from the top-left corner to the bottom-right corner that maximizes the sum of the numbers on it. You can only move rightwards or downwards.\\ 
\\
Table: \\
85 93 45 79 49 \\
28 12 37 57 76 \\
3 22 37 55 68 \\
26 2 57 7 100 \\
87 11 12 67 89 \\
\\
    \end{tcolorbox}
    \caption{Example Prompt for the dynamic programming task.}
    \label{fig:app_dp_prompt}
\end{figure*}

\begin{figure*}[htbp]
    \centering
    \begin{tcolorbox}[
        title=Full CoT example,
        coltitle=white,
        colbacktitle=black,
        colback=gray!10,
        colframe=gray!50,
        width=0.95\textwidth,
    ]
Find a path in the given table from the top-left corner to the bottom-right corner that maximizes the sum of the numbers on it. You can only move rightwards or downwards.\\
\\
Table:\\
15 5 59 62 22\\
41 61 7 12 27\\
98 60 34 94 24\\
45 40 12 77 11\\
56 94 46 34 45\\
\\
<tool\_call>15 20 79 141 163\\
56 117 124 153 190\\
154 214 248 342 366\\
199 254 266 419 430\\
255 349 395 453 498</tool\_call>\\
\\
Result: 498
    \end{tcolorbox}
    \caption{Example CoT for the dynamic programming task.}
    \label{fig:app_dp_cot}
\end{figure*}

\section{Main Experiment Settings}
\label{sec:appendix_main}
For all of our experiments, we use Qwen-2.5-1.5B~\cite{yang2024qwen2} from the huggingface model hub as the base model.
On each task, we finetune the model on the training set and then evaluate the model on the test set of the corresponding prompt type.
We use the full-parameter supervised finetuning setting and do not use parameter-efficient training techniques.

During training, we use the AdamW optimizer with a learning rate of $1e-5$.
The weight decay is set to 0 and the gradient clipping threshold is set to 1.
We train the model for 1 epoch with a training batch size of 4 by default.
For small datasets like $1 \times 2$ digit multiplication, we change the epoch to 10 to ensure convergence.

The models on multiplication problems are trained under BFloat16 precision, while models on DP problems are trained under Float32 precision.

During evaluation, we evaluate with a batch size of 1.
We only check the correctness of the final result, and do not check the values in CoT.

\section{Latent Experiment Settings}
\label{sec:appendix_latent}
The hyperparameters in latent experiments are the same as the main experiment.
For convenience, we use \texttt{<|fim\_middle|>} as the latent token \texttt{<LAT>}.

In multiplication problems, the dimension of latent embeddings is set to 20 (10 for digit results and 10 for carry results).
In dynamic programming problems, the dimension of latent embeddings is set to 50 to store values no larger than 100,000.
The latent projection module $P_{in}$ and the latent output head $P_{out}$ are trained with the backbone model with the same learning rate.
We simply add the latent loss $\mathcal{L}_{lat}$ with the original LM head loss $\mathcal{L}_{lm}$ as the final loss $\mathcal{L} = \mathcal{L}_{lat} + \mathcal{L}_{lm}$.

Figure \ref{fig:app_multi_latent} shows an example of latent CoT in multiplication problems, and Figure \ref{fig:app_dp_latent} shows an example of latent CoT in dynamic programming problems.

\begin{figure*}[htbp]
    \centering
    \begin{tcolorbox}[
        title=Latent CoT example,
        coltitle=white,
        colbacktitle=black,
        colback=gray!10,
        colframe=gray!50,
        width=0.95\textwidth,
    ]
8493*8877=<tool\_call>8493*7\\
<|fim\_middle|><|fim\_middle|><|fim\_middle|><|fim\_middle|>|59451\\
8493*70
<|fim\_middle|><|fim\_middle|><|fim\_middle|><|fim\_middle|>|594510\\
8493*800\\
<|fim\_middle|><|fim\_middle|><|fim\_middle|><|fim\_middle|>|6794400\\
8493*8000\\
<|fim\_middle|><|fim\_middle|><|fim\_middle|><|fim\_middle|>|67944000\\
\\
59451+594510+6794400+67944000\\
59451+594510+6794400+67944000=653961+6794400+67944000\\
653961+6794400+67944000=7448361+67944000\\
7448361+67944000=75392361\\
\\
8493*8877=75392361</tool\_call>\\
\\
Result: 75392361
    \end{tcolorbox}
    \caption{Example latent CoT for the multi-digit multiplication task.}
    \label{fig:app_multi_latent}
\end{figure*}

\begin{figure*}[htbp]
    \centering
    \begin{tcolorbox}[
        title=Latent CoT example,
        coltitle=white,
        colbacktitle=black,
        colback=gray!10,
        colframe=gray!50,
        width=0.95\textwidth,
    ]
Find a path in the given table from the top-left corner to the bottom-right corner that maximizes the sum of the numbers on it. You can only move rightwards or downwards.\\
\\
Table:\\
15 5 59 62 22\\
41 61 7 12 27\\
98 60 34 94 24\\
45 40 12 77 11\\
56 94 46 34 45\\
\\
<tool\_call><|fim\_middle|><|fim\_middle|><|fim\_middle|><|fim\_middle|><|fim\_middle|>\\
<|fim\_middle|><|fim\_middle|><|fim\_middle|><|fim\_middle|><|fim\_middle|>\\
<|fim\_middle|><|fim\_middle|><|fim\_middle|><|fim\_middle|><|fim\_middle|>\\
<|fim\_middle|><|fim\_middle|><|fim\_middle|><|fim\_middle|><|fim\_middle|>\\
<|fim\_middle|><|fim\_middle|><|fim\_middle|><|fim\_middle|><|fim\_middle|></tool\_call>\\
\\
Result: 498
    \end{tcolorbox}
    \caption{Example latent CoT for the dynamic programming task.}
    \label{fig:app_dp_latent}
\end{figure*}

\section{Intervention Experiment Details}
\label{sec:appendix_intervention}
In the intervention experiments, we randomly substitute a number value in the CoT generated by trained models on the test set.
The interventions are performed on the full CoT texts.
The substituted number has the same digit length as the original number, but with a different value.
To prevent outlier values, we keep the first digit to be the same as the original number when substituting numbers with 2 or more digits.

We choose the number to substitute within the following range:
\paragraph{Multiplication}
\begin{itemize}
    \item $x$ or $y$ in ``digit $x$, carry $y$'' statements;
    \item A random number $x$ in ``Add up partial results:'' statements;
    \item The first partial result $x$ in ``$a_1 + \ldots + a_n = x + \ldots $ statements;
    \item The result $x$ in the ``The final result is: $\ldots = x$'' statement.
\end{itemize}

\paragraph{Dynamic programming}
A random intermediate value in the CoT.

After intervention, we truncate all tokens after the intervened value, and feed the partial CoT into trained models to complement the full CoT and get the final answer.

The detailed breakdown of errors in multiplication problems is shown in Table \ref{tab:app_intervention_breakdown} (1 entry with deformed CoT is excluded):
\begin{table}[htbp]
    \centering
    \begin{tabular}{c|c}
    \hline
    Type & Count \\
    \hline
    Total & 9999 \\
    Success & 7383 \\
    Error & 2616 \\
    Addition error & 767 \\
    Reconstruct error & 496 \\
    Shortcut error & 1291 \\
    Copy error & 6 \\
    Misc error & 56 \\
    \hline
    \end{tabular}
    \caption{Intervention error breakdown in multiplication problems.}
    \label{tab:app_intervention_breakdown}
\end{table}

\section{Probing Experiment Details}
\label{sec:appendix_probe}
In the probing experiments, we probe on latent CoT for simplicity.
We first collect hidden states of LLMs on different layers, and then train the probe classifiers.
The training set of hidden states is collected by running the trained model on the original training set, and so is the test set.

We use a learning rate of $1e-3$ and a gradient clipping threshold of $1$.
We train the probe classifiers for 4 epochs with a training batch size of 32, and an evaluate batch size of 64.

\end{document}